# Multi-layer Kernel Ridge Regression for One-class Classification

Chandan Gautam[a,*], Aruna Tiwari[a], Sundaram Suresh[b], Alexandros Iosifidis[c]

[a]*Indian Institute of Technology Indore, Indore, Simrol,India*
[b]*School of Electrical and Electronic Engineering, Nanyang Technological University Singapore*
[c]*Aarhus University, Denmark*

**Abstract**

In this paper, a multi-layer architecture (in a hierarchical fashion) by stacking various Kernel Ridge Regression (*KRR*) based Auto-Encoder for one-class classification is proposed and is referred as MKOC. MKOC has many layers of Auto-Encoders to project the input features into new feature space and the last layer is regression based one class classifier. The Auto-Encoders use an unsupervised approach of learning and the final layer uses semi-supervised (trained by only positive samples) approach of learning. The proposed MKOC is experimentally evaluated on 15 publicly available benchmark datasets. Experimental results verify the effectiveness of the proposed approach over 11 existing state-of-the-art kernel-based one-class classifiers. Friedman test is also performed to verify the statistical significance of the claim of the superiority of the proposed one-class classifiers over the existing state-of-the-art methods.

## 1. Introduction

**O**ne-**c**lass **C**lassification (OCC) has been widely used for outlier, novelty, fault, and intrusion detection [1, 2, 3, 4] by researchers from different disciplines. In OCC problems, samples of the class of interest (i.e., positive samples) are available while negative samples are very rare or costly to collect [5, 6, 7, 8, 9], thus making the application of multi-class models problematic. Various one-class classifiers [1, 2] have been proposed based on the regression model, the clustering model etc. One-class classification methods available in the literature can be divided into two broad categories viz., non-kernel-based and kernel-based methods. Various non-kernel-based one-class classifiers are principal component analysis based data descriptor[1] [5], angle-based outlier factor data description [10], K-means data

---

description [5], self-organizing map data description [5], Auto-Encoder data descriptor [11] etc. Whereas, the kernel-based one-class classifier approaches are support vector data description [12], one-class support vector machine[13], kernel principal component analysis based data description[14] etc. However, kernel-based methods have been shown to outperform non-kernel-based methods in the literature [1, 5]. Despite this fact, these kernel-based methods involve the solution of a quadratic optimization problem, which is computationally expensive. Apart of these kernel-based methods, *KRR*-based models [15] optimize the problem rapidly in a non-iterative way by solving a linear systems. Therefore, *KRR*-based models [15, 16, 17, 18, 19] have received quite attention by researchers for solving various types of problems viz., regression, binary, multi-class etc.

In recent years, various *KRR*-based one-class classifiers have been developed and exhibited better performance compared to various state-of-the-art one-class classifiers. Overall, the *KRR*-based one-class classifiers can be divided into two types, namely, (i) without Graph-Embedding (ii) with Graph-Embedding. For 'without Graph-Embedding', two types of architectures have been explored for OCC. One is *KRR*-based single output node architecture [20], and other is *KRR*-based Auto-Encoder architecture [21]. For 'with Graph-Embedding', Iosifidis et al.[22] presented local and global structure. Different types of Laplacian Graphs are employed for local (i.e., Local Linear Embedding, **L**aplacian **E**igenmaps etc.) and global (linear discriminant analysis and clustering-based discriminant analysis etc.) Graph-embedding. Later, global variance-based Graph-Embedding has been extended in order to exploit class variance and sub-class variance information for face verification task by Mygdalis et al.[23]. All the above-mentioned *KRR*-based one-class classifiers employ a single-layered architectures.

Over the last decade, stacked Auto-encoder based multi-layer architectures have received quite attention by researchers for multi-class or binary class classification tasks [24, 25]. Such architectures can lead to better representation learning [26, 27] and also used in dimensionality reduction [28, 29, 30]. High-level feature representations obtained by using stacked Auto-Encoder also helps in improving the performance of the traditional classifiers [31]. Moreover, this concept is also extended for kernel-based learning [32, 33]. Recently, Wong et al. [34] explores *KRR*-based representation learning for the multi-class classification task. Inspired by the advantage of multi-layer architecture, this paper explores the possibility of *KRR*-based representation learning for the one-class classification task.

In this paper, we propose a multi-layer architecture by stacking various *KRR*-based Auto-Encoder (trained using unsupervised learning) in a hierarchical manner for one-class classification task. After stacking several Auto-Encoder layers in a hierarchical manner, data are represented in a new feature space in which regression-based one-class classification is employed in the final layer. Overall architecture is referred as **M**ulti-layer **K**RR-based architecture for **O**ne-class **C**lassification (***MKOC***). The multiple layers exploit the idea of successive nonlinear data mappings and hence capture the relationship effectively. During the training process of the Auto-Encoder, it simultaneously enhances the data reconstruction and data representation ability of the classifier. Output of the stacked Auto-Encoder is approximated to any real number and set a threshold for deciding whether any sample is outlier or not. Two types of threshold deciding criteria are discussed so far in this paper.

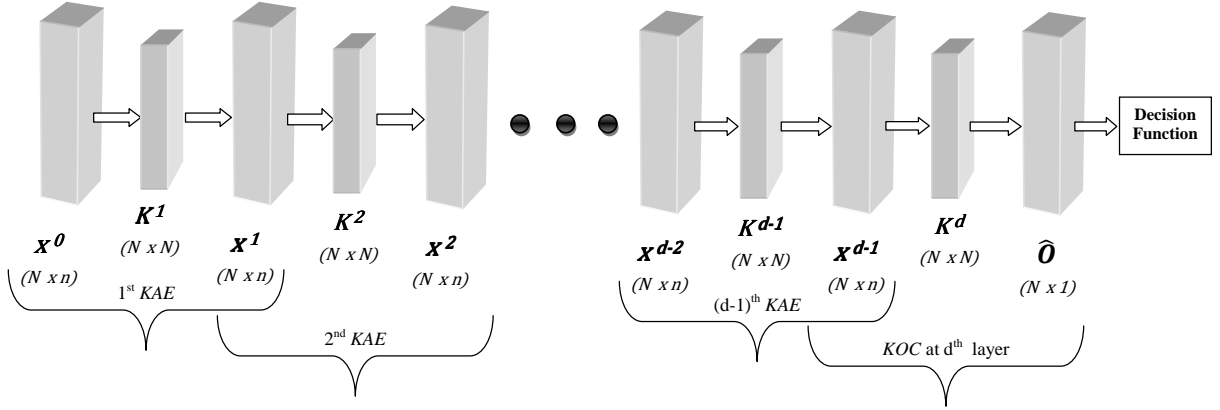

Figure 1: Schematic Diagram of **MKOC**

Further, the *MKOC* performance is evaluated using 15 benchmark datasets and its performance is compared with 11 state-of-the-art kernel-based methods available in the literature. Finally, a Friedman test [35] is conducted to verify the statistical significance of the experimental outcomes of *MKOC* classifier and it rejects the null hypothesis with 95% confidence level.

The rest of the paper is organized as follows. Section 2 describes the MKOC in detail. Performance evaluation is provided in Section 3. Finally, Section 4 concludes our work.

## 2. Multi-Layer KRR based Once-Class Classifier

In this section, a **M**ulti-layer **K**RR-based architecture for **O**ne-class **C**lassification (*MKOC*) is described. The proposed multi-layer architecture is constructed by stacking various *KRR*-based Auto-Encoders (*KAEs*), followed by a *KRR*-based one-class classifier (KOC), as shown in Fig. 1. These stacked Auto-Encoders are employed for defining the successive data representation. In the $1^{st}$ *KAE* of this figure, input training matrix is denoted by $X = X^0 = \{x_i^0\}$, where $x_i^0 = [x_{i1}^0, x_{i2}^0, ..., x_{in}^0]$, $i = 1, 2, ..., N$, is the $n$-dimensional input vector of the $i^{th}$ training sample. Let us assume that there are $d$ layers in the proposed architecture, i.e., $h = 1, 2, ..., d$. Output of the $h^{th}$ layer is passed as input to the $(h+1)^{th}$ layer. Let us denote output at $h^{th}$ layer of Auto-Encoder, $X^h = \{x_i^h\}$, where $x_i^h = [x_{i1}^h, x_{i2}^h, ..., x_{in}^h]$, $i = 1, 2, ..., N$. $X^h$ corresponds to the output of the $h^{th}$ Auto-Encoder and the input of the $(h+1)^{th}$ Auto-Encoder. Each of the Auto-Encoders involves a data mapping using function $\phi(.)$, mapping $X^{h-1}$ to $\phi^h = \phi(X^{h-1})$. $\phi(.)$ corresponds to a mapping of $X^{h-1}$ to the corresponding kernel space $K^h = (\Phi^h)^T \Phi^h$. Here, $\Phi^h = \left[\phi_1^h, \phi_2^h, ..., \phi_N^h\right]$. The data representation obtained by calculating the output of the $(d-1)^{th}$ Auto-Encoder in the architecture is passed to the $d^{th}$ layer for OCC using *KOC*. Two types of training errors are generated by *MKOC*. One is generated by the Auto-Encoder until $d-1$ layers and denoted as an error matrix $E^h = \{e_i^h\}$, where $i = 1, 2, ..., N$ and $h = 1, 2, ..., (d-1)$. And other is generated by the one-class classifier at $d^{th}$ layer which is denoted as an error vector $E^d = \{e_i^d\}$, where $i = 1, 2, ..., N$. The overall architecture of *MKOC* is formed by two processing steps.

In **first step**, $(d-1)$ *KAE*s are trained, each defining a pair $(X^h, \beta_a^h)$, and stacked in a hierarchical manner. Here, $\beta_a^h$ denotes weight matrix of the $h^{th}$ Auto-Encoder. A *KAE* minimizes the following criterion, which involves a non-linear[2] feature mapping $X^{h-1} \rightarrow \Phi^h$:

$$\text{Minimize}: \pounds_{KAE} = \frac{1}{2}\left\|\beta_a^h\right\|^2 + \frac{C}{2}\sum_{i=1}^{N}\left\|e_i^h\right\|_2^2$$
$$\text{Subject to}: (\beta_a^h)^T \phi_i^h = x_i^{h-1} - e_i^h, \ i = 1, 2, ..., N, \quad (1)$$

where $C$ is a regularization parameter, and $e_i^h$ is a training error vector corresponding to the $i^{th}$ training sample at $h^{th}$ layer. Based on the Representer Theorem [36], we express $\beta_a^h$ as a linear combination of the training data representation $\Phi^h$ and a reconstruction weight matrix $W_a^h$:

$$\beta_a^h = \Phi^h W_a^h. \quad (2)$$

Hence, by using Representer Theorem [36], minimization criterion in (1) is reformulated as follows:

$$\text{Minimize}: \pounds_{KAE} = \frac{1}{2}Tr\left((W_a^h)^T(\Phi^h)^T \Phi^h W_a^h\right) + \frac{C}{2}\sum_{i=1}^{N}\left\|e_i^h\right\|_2^2,$$
$$\text{Subject to}: (W_a^h)^T(\phi_i^h)^T \phi_i^h = x_i^{h-1} - e_i^h, \ i = 1, 2, ..., N. \quad (3)$$

By further substitution of $K^h = (\Phi^h)^T \Phi^h$, where $k_i^h \subseteq K^h$ is formed by the elements $k_{ij}^h = (\phi_i^h)^T \phi_j^h$, the criterion in (3) can be written as:

$$\text{Minimize}: \pounds_{KAE} = \frac{1}{2}Tr\left((W_a^h)^T K^h W_a^h\right) + \frac{C}{2}\sum_{i=1}^{N}\left\|e_i^h\right\|_2^2,$$
$$\text{Subject to}: (W_a^h)^T k_i^h = x_i^{h-1} - e_i^h, \ i = 1, 2, ..., N. \quad (4)$$

---
[2] Linear case can be easily derived from (1) by substituting $\phi_i^h \rightarrow x_i^{h-1}$

The Lagrangian relaxation of (4) is shown below in (5):

$$\pounds_{KAE} = \frac{1}{2}Tr\left((W_a^h)^T K^h W_a^h\right) + \frac{C}{2}\sum_{i=1}^{N}\|e_i^h\|_2^2 \\ - \sum_{i=1}^{N}\alpha_i^h((W_a^h)^T k_i^h - x_i^{h-1} + e_i^h) \tag{5}$$

where $\alpha = \{\alpha_i^h\}, i = 1, 2 \ldots N$, is a Lagrangian multiplier. In order to optimize $\pounds_{LKAE}$, we compute its derivatives as follows:

$$\frac{\partial \pounds_{KAE}}{\partial W_a^h} = 0 \Rightarrow W_a^h = \alpha \tag{6}$$

$$\frac{\partial \pounds_{KAE}}{\partial e_i^h} = 0 \Rightarrow E^h = \frac{1}{C}\alpha \tag{7}$$

$$\frac{\partial \pounds_{KAE}}{\partial \alpha_i^h} = 0 \Rightarrow (W_a^h)^T K^h = X^{h-1} - E^h \tag{8}$$

The matrix $W_a^h$ is obtained by substituting (7) and (8) into (6), and is given by:

$$W_a^h = \left(K^h + \frac{I}{C}\right)^{-1} (X^{h-1})^T \tag{9}$$

Now, $\beta_a^h$ can be derived by substituting (9) into (2):

$$\beta_a^h = \Phi^h \left(K^h + \frac{I}{C}\right)^{-1} (X^{h-1})^T \tag{10}$$

Hence, the transformed data $X^h$ by the $h^{th}$ KAE can be obtained as follows:

$$X^h = (\Phi^h)^T \beta_a^h = (\Phi^h)^T \Phi^h W_a^h = K^h (W_a^h)^T \tag{11}$$

where $K^h \in \mathbb{R}^{N \times N}$ is the kernel matrix for the $h^{th}$ layer. After mapping the training data through the $(d-1)$ successive *KAE*s **in the first step**, the training data representations defined by the outputs of the $(d-1)^{th}$ *KAE* are used in order to train a *KOC* **in the second step**. *KOC* involves a nonlinear feature mapping $X^{d-1} \to \Phi^d$ and is trained by solving the following optimization problem:

$$\text{Minimize} : \pounds_{MKOC^d} = \frac{1}{2} \left\| \beta_o^d \right\|^2 + \frac{C}{2} \sum_{i=1}^{N} \left\| e_i^d \right\|_2^2$$

$$\text{Subject to} : (\beta_o^d)^T \phi_i^d = r - e_i^d, \; i = 1, 2, ..., N \tag{12}$$

where $e_i^d$ is training error corresponding to $i^{th}$ training sample and $\beta_o^d$ denotes weight vector at $d^{th}$ layer. By using Representer Theorem [36], $\beta_o^d$ is expressed as a linear combination of the training data representation $\Phi^d$ and reconstruction **weight vector** $W_o^d$:

$$\beta_o^d = \Phi^d W_o^d. \tag{13}$$

Hence, the minimization criterion in (12) is reformulated to the following:

$$\text{Minimize} : \pounds_{MKOC^d} = \frac{1}{2}(W_o^d)^T (\Phi^d)^T \Phi^d W_o^d + \frac{C}{2} \sum_{i=1}^{N} \left\| e_i^d \right\|_2^2$$

$$\text{Subject to} : (W_o^d)^T (\phi_i^d)^T \phi_i^d = r - e_i^d, \; i = 1, 2, ..., N \tag{14}$$

In addition, by substituting $K^d = (\Phi^d)^T \Phi^d$, where $k_i^d \subseteq K^d$, the optimization problem in (14) can be reformulated as follows:

$$\text{Minimize} : \pounds_{MKOC^d} = \frac{1}{2}(W_o^d)^T K^d W_o^d + \frac{C}{2} \sum_{i=1}^{N} \left\| e_i^d \right\|_2^2,$$

$$\text{Subject to} : (W_o^d)^T k_i^d = r - e_i^d, \; i = 1, 2, ..., N. \tag{15}$$

The Lagrangian relaxation of (15) is shown below in (16):

$$\pounds_{MKOC^d} = \frac{1}{2}(W_o^d)^T K^d W_o^d + \frac{C}{2} \sum_{i=1}^{N} \left\| e_i^d \right\|_2^2$$

$$- \sum_{i=1}^{N} \alpha_i^d ((W_o^d)^T k_i^d - r + e_i^d) \tag{16}$$

where $\alpha = \{\alpha_i^d\}, i = 1, 2 \ldots N$, is a Lagrangian multiplier. In order to optimize $\pounds_{MKOC^d}$, we compute its derivatives as follows:

$$\frac{\partial \pounds_{MKOC^d}}{\partial W_o^d} = 0 \Rightarrow W_o^d = \alpha \tag{17}$$

$$\frac{\partial \pounds_{MKOC^d}}{\partial e_i^d} = 0 \Rightarrow E^d = \frac{1}{C}\alpha \tag{18}$$

$$\frac{\partial \pounds_{MKOC^d}}{\partial \alpha_i^d} = 0 \Rightarrow (W_o^d)^T K^d = r - E^d \tag{19}$$

The matrix $W_o^d$ of $d^{th}$ layer is, thus, obtained by substituting (18) and (19) into (17), and is given by:

$$W_o^d = \left(K^d + \frac{I}{C}\right)^{-1} r \tag{20}$$

$\beta_o^d$ can be derived by substituting (20) in (13):

$$\beta_o^d = \Phi^d \left(K^d + \frac{I}{C}\right)^{-1} r, \tag{21}$$

where $r$ is a vector having all elements equal to $r$. Since the value $r$ can be arbitrary, we set it equal to $r = 1$.

The predicted output of the final layer (i.e., $d^{th}$ layer) of the multi-layer architecture for training samples can be calculated as follows:

$$\widehat{O} = (\Phi^d)^T \beta_o^d = (\Phi^d)^T \Phi^d W_o^d = K^d (W_o^d)^T \tag{22}$$

where $\widehat{O}$ is the predicted output for training data.

After completing the training process, a threshold is required to decide whether any sample is an outlier or not which is discussed in the next subsection.

*2.1. Decision Function*

Two types of thresholds namely, $\theta 1$ and $\theta 2$, are employed with the proposed method, which are determined as follows:

1. **For $\theta 1$:**

    (i) Calculate distance between the predicted value of the $i^{th}$ training sample and $r$, and store in a vector $d$ as follows:

    $$d(i) = \left|\widehat{O}_i - r\right| \tag{23}$$

    (ii) After storing all distances in $d$ as per (23), sort these distances in decreasing order and denoted by a vector $d_{dec}$. Further, reject few percent of training samples based on the deviation. Most deviated samples are rejected first because they are most probably far from the distribution of the target data. The threshold is decided based on these deviations as follows:

    $$\theta 1 = d_{dec}(\lfloor \eta * N \rfloor) \tag{24}$$

where $0 < \eta \leq 1$ is the fraction of rejection of training samples for deciding threshold value. $N$ is the number of training samples and $\lfloor \ \rfloor$ denotes floor operation.

2. **For $\theta2$:** Select threshold ($\theta2$) as a small fraction of the mean of the predicted output:

$$\theta2 = (\lfloor \eta * \text{mean}(\widehat{O}) \rfloor) \tag{25}$$

where $0 < \eta \leq 1$ is the fraction of rejection for deciding threshold value.

Hence, a threshold value can be determined by above procedures. Afterwards, during testing, a test vector $x_p$ is fed to the trained multi-layer architecture and its output $\widehat{O}_p$ is obtained. Further, compute $\widehat{d}$ for both types of threshold as follows:

For $\theta1$, calculate the distance ($\widehat{d}$) between the predicted value $\widehat{O}_p$ of the $p^{th}$ testing sample and $r$:

$$\widehat{d} = \left| \widehat{O}_p - r \right| \tag{26}$$

For $\theta2$, calculate the distance ($\widehat{d}$) between the predicted value $\widehat{O}_p$ of the $p^{th}$ testing sample and mean of the predicted values obtained after training as follows:

$$\widehat{d} = \left| \widehat{O}_p - \text{mean}(\widehat{O}) \right| \tag{27}$$

---

**Algorithm 1** Multi-layer KRR-based architecture for OCC: $MKOC$
---
**Input:** Training set $X$, regularization parameter ($C$), kernel function ($\Phi$), number of layers ($d$)
**Output:** Whether incoming sample is target or outlier
1: Initially, $X^0 = X$
2: **for** $h = 1$ to $d$ **do**
3:    **if** $h < d$ **then**
4:       **First Phase**: $KAE$s are stacked from first to $(d-1)^{th}$ layer in the hierarchical fashion and transform the input samples.
5:       Train the $KAE$ as per Eq. (4).
6:       Transformed output $X^h$ for the input $X^{h-1}$ is computed to pass as the input to the next layer in the hierarchy.
7:    **else**
8:       **Second Phase**: Final layer, i.e., $d^{th}$ layer for one-class classification.
9:       Output of $(d-1)^{th}$ Auto-Encoder is passed as an input to one-class classifier at $d^{th}$ layer.
10:      Train the $d^{th}$ layer by $MKOC^d$ as per Eq. (15).
11:    **end if**
12: **end for**
13: Compute a threshold either $\theta1$ (Eq. (24)) or $\theta2$ (Eq. (25)).
14: At final step, whether a new input is outlier or not, decides based on the rule discussed in Eq. (28).

---

Finally, $x_p$ is classified based on the following rule:

$$\text{If } \widehat{d} \leq \text{Threshold}, \quad x_p \text{ belongs to normal class} \tag{28}$$
$$\text{Otherwise}, \quad x_p \text{ is an outlier}$$

The overall processing steps followed by *MKOC* are described in Algorithm 1.

The above-described multi-layer OCC architecture creates two variants of *MKOC* using two types of threshold criteria (viz., $\theta 1$ and $\theta 2$), i.e., $MKOC\_\theta 1$ and $MKOC\_\theta 2$.

## 3. Performance Evaluation

In this section, experiments are conducted to evaluate the performance of the proposed MKOC over 15 data sets. These datasets are obtained from University of California Irvine (UCI) repository [37] and were originally generated for the binary or multi-class classification task. For our experiments, we have made it compatible with OCC task in the following ways. If a dataset has two classes then, alternately, we use each of the classes in the binary dataset as the target class and the remaining one as outlier. If a dataset has more than two classes then we use one of the classes in the dataset as the target class and the remaining ones as outliers. In this way, we construct 15 one-class datasets from 8 multi-class datasets. Description of these datasets can be found in Table 1. These 15 datasets can be divided into 3 category viz., 6 financial, 6 medical and 3 miscellaneous datasets. Many of the datasets are slightly imbalanced. Class imbalance ratio of both of the classes are approximately 1 : 2 in case of 7 datasets viz., German(1), German(2), Pima(1), Pima(2), Glass(1), Glass(2), and Iris. All experiments on these datasets are carried out with MATLAB 2016a on Windows 7 (Intel Xeon 3 GHz processor, 64 GB RAM) environment.

Total 11 existing kernel-based one-class classifiers are employed for comparison purpose, which can be categorized as follows:

(i) Support Vector Machine (*SVM*) based: **O**ne-**c**lass **SVM** (*OCSVM*) [13], **S**upport **V**ector **D**ata **D**escription (*SVDD*) [12]

(ii) *KRR*-based:

  (a) Without Graph-Embedding: **K**RR-based **OCC** (*KOC*) [20] and **K**RR-based **A**uto-**E**ncoder model for **OCC** (*AEKOC*) [21]

  (b) With Graph-Embedding: Two types of Graph-Embedding, i.e., Local and Global, have been explored in the literature. **L**ocal and **G**lobal Graph-Embedding with *KOC* are named as *LKOC*-X [22] and *GKOC*-X [22, 23], respectively. Here, X can be any Laplacian Graph with local or global Graph-embedding. For local, two types of Graphs are explored viz., **L**ocal **L**inear **E**mbedding (*LLE*) and **L**aplacian **E**igenmaps (*LE*). For global, four types of Graphs are explored viz., **L**inear Discriminant Analysis (*LDA*), **C**lustering-based L**DA** (*CDA*), class variance (*CV*), and sub-class variance (*SV*). Hence, final six variants are generated namely, *LKOC-LE*[22], *LKOC-LLE* [22], *GKOC-LDA* [22], *GKOC-CDA*[22], *GKOC-CV*[23] and *GKOC-SV*[23].

(iii) Principal Component Analysis (*PCA*) based: **K**ernel **PCA** (*KPCA*)[14].

All one-class classifiers are implemented and tested in the same environment. *OCSVM* is implemented using LIBSVM library [38]. *SVDD* is implemented by using DD Toolbox [39].

Codes of all *KRR*-based one-class classifiers were provided by the authors of the corresponding papers. The implementations of *KPCA*[14] and *AEKOC*[21] are obtained from the links given in the paper (links are made available at the reference of the corresponding paper).

Table 1: Datasets Description

| S. No. | Name | #Targets | #Outliers | #Features | #samples |
|---|---|---|---|---|---|
| **Financial Credit Approval Datasets** | | | | | |
| 1 | German(1) | 700 | 300 | 24 | 1000 |
| 2 | German(2) | 300 | 700 | 24 | 1000 |
| 3 | Australia(1) | 307 | 383 | 14 | 690 |
| 4 | Australia(2) | 383 | 307 | 14 | 690 |
| 5 | Japan(1) | 294 | 357 | 15 | 651 |
| 6 | Japan(2) | 357 | 294 | 15 | 651 |
| **Medical Disease Datasets** | | | | | |
| 7 | Heart(1) | 160 | 137 | 13 | 297 |
| 8 | Heart(2) | 137 | 160 | 13 | 297 |
| 9 | Pima(1) | 500 | 268 | 8 | 768 |
| 10 | Pima(2) | 268 | 500 | 8 | 768 |
| 11 | Bupa(1) | 145 | 200 | 6 | 345 |
| 12 | Bupa(2) | 200 | 145 | 6 | 345 |
| **Miscellaneous Datasets** | | | | | |
| 13 | Glass(1) | 76 | 138 | 9 | 214 |
| 14 | Glass(2) | 138 | 76 | 9 | 214 |
| 15 | Iris | 50 | 100 | 4 | 150 |

For all of the kernel-based methods, Radial Basis Function (RBF) kernel is employed as shown below,

$$\kappa(x_i, x_j) = exp\left(-\frac{\|x_i - x_j\|_2^2}{2\sigma^2}\right) \quad (29)$$

where $\sigma$ is calculated as the mean Euclidean distance between training vectors in the corresponding feature space. For the proposed multi-layer architecture (*MKOC*), we have used maximum $d = 5$ layers and the value of $\sigma^h$ is calculated at each $h^{th}$ layer independently using the training data representations $X^{h-1}$. At each layer, regularization parameter is selected from the range of $\{2^{-3}, \ldots, 2^3\}$. The classifiers which exploit graphs, i.e., *LKOC*-X, *GKOC*-X, and *GKOC*-XX, have two regularization parameters, which are selected based on the cross-validation using values $2^l$, where $l = \{-3, ..., 3\}$. For graph encoding subclass information in *GKOC*-XX, the number of subclasses is selected from the range $\{2, 3, ..., 20\}$. For the *KOC* and *AEKOC* methods, regularization parameter is selected from the range $\{2^{-3}, \ldots, 2^3\}$. For *KPCA* based OCC, the percentage of the preserved variance is selected from the range $[85, 90, 95]$. The fraction of rejection ($\eta$) of outliers during threshold selection is set equal to 0.05 for all methods.

### 3.1. Performance Evaluation Criteria

Geometric mean ($\eta_g$) is computed in the experiment for evaluating the performance of each of the classifiers and is calculated as

$$\eta_g = \sqrt{\text{Precision} * \text{Recall}} \quad (30)$$

In all our experiments, 5-fold cross-validation (CV) procedure is used and the average Gmean value (along with the corresponding standard deviation ($\Delta$)) over 5-fold CV are reported in the results. $\eta_g$ values of all the classifiers are further analyzed by using mean of all Gmeans ($\eta_m$) and percentage of the maximum Gmean ($\eta_p$). $\eta_m$ is computed by taking average of all Gmeans obtained by a classifier over all datasets. $\eta_p$ is computed as follows [40]:

$$\eta_p = \frac{\sum_{i=1}^{\text{no. of datasets}} \left(\frac{\text{Gmean of classifier for } i^{th} \text{ dataset}}{\text{Maximum Gmean achieved for } i^{th} \text{ dataset}} \times 100\right)}{\text{Number of datasets}} \quad (31)$$

Moreover, Friedman testing is performed to verify the statistical significance of the obtained results. To this end, similar to [40], we also compute Friedman Rank ($\eta_f$)[35].

### 3.2. Performance Comparison

The Gmean ($\eta_g$) of the 13 kernel-based methods over 15 datasets are provided in Tables 2 to 4 for financial, medical, and miscellaneous datasets, respectively. Best $\eta_g$ per dataset is displayed in boldface in these Tables. Out of 6 **financial** credit approval datasets, proposed multi-layer one-class classifier performs better for 4 datasets. In case of Australian(1) dataset, $MKOC\_\theta1$ and $MKOC\_\theta2$ yield significantly (>4%) better results compared to all of the methods presented in Table 2 except $AEKOC$. However, both exhibit comparable performance to $AEKOC$. Out of 6 **medical** datasets in Table 3, proposed multi-layer one-class classifier performs better for 3 datasets and yields comparable $\eta_g$ for rest of the 3 datasets. Moreover, $MKOC\_\theta1$ and $MKOC\_\theta2$ perform better for 1 and 2 datasets, respectively. Further, among 3 **miscellaneous** datasets, $MKOC\_\theta1$ yields better results for Glass(1) and Iris datasets.

Table 2: Performance in term of $\eta_g \pm \Delta$ (%) over 5-folds and 5 runs for financial datasets

| One-class Classifier | German(1) | German(2) | Australia(1) | Australia(2) | Japan(1) | Japan(2) |
|---|---|---|---|---|---|---|
| KPCA [14] | 80.77±0.07 | 49.75±0.28 | 63.69±0.29 | 73.06±0.18 | 64.09±0.29 | 72.29±0.29 |
| OCSVM [13] | 80.34±0.33 | 52.8±0.86 | 66.08±0.6 | 76.59±0.44 | 71.45±0.38 | 75.78±0.29 |
| SVDD [12] | 81.1±0.34 | 52.77±0.82 | 65.55±0.47 | 76.78±0.22 | 70.15±0.4 | 76.58±0.28 |
| KOC [20] | 73.17±0.26 | 53.41±0.3 | 65.07±0.68 | 74.21±1.12 | 67.33±1.24 | 73.48±1.62 |
| AEKOC [21] | 74.04±0.29 | 51.57±0.37 | **72.88**±0.66 | 77.96±0.55 | **76.23**±0.46 | 78.27±0.51 |
| GKOC-LDA [22] | 72.53±0.74 | 52.94±0.24 | 64.98±1.09 | 73.71±1.21 | 67.12±1.19 | 72.73±1.83 |
| LKOC-LE [22] | 72.75±0.5 | 53.06±0.34 | 65.09±0.82 | 74.03±1.26 | 67.24±1.2 | 73.15±1.63 |
| LKOC-LLE [22] | 70.86±0.52 | 52.51±0.7 | 62.95±0.64 | 70.72±0.68 | 64.97±0.39 | 69.79±1.52 |
| GKOC-CDA [22] | 72.6±0.66 | 52.9±0.15 | 67.48±1.67 | 73.74±1.2 | 74.58±2.56 | 72.75±1.8 |
| GKOC-CV[23] | 81.42±0.17 | 53.06±0.08 | 63.21±0.4 | 73.67±0.51 | 63.91±0.43 | 73.56±0.36 |
| GKOC-SV [23] | 79.39±1.63 | 53.49±0.39 | 63.9±0.67 | 74.6±1.16 | 66.17±0.69 | 73.96±0.92 |
| MKOC_$\theta$1 | 74.1±0.86 | **54.46**±0.11 | 72.12±0.76 | 79.89±0.75 | 74.45±0.59 | 80.14±0.76 |
| MKOC_$\theta$2 | **82.79**±0.3 | 49.59±0.76 | 72.43±0.52 | **80.31**±0.45 | 74.66±0.83 | **80.55**±0.28 |

Table 3: Performance in term of $\eta_g \pm \Delta$ (%) over 5-folds and 5 runs for medical datasets

| One-class Classifier | Heart(1) | Heart(2) | Pima(1) | Pima(2) | Bupa(1) | Bupa(2) |
|---|---|---|---|---|---|---|
| KPCA [14] | 70.42±0.22 | 63.5±0.78 | 77.98±0.18 | 57.05±0.4 | **62.91**±0.4 | 74.28±0.59 |
| OCSVM [13] | 72.91±0.55 | 64.9±1.4 | 79.18±0.19 | 56.59±0.47 | 60.64±1.3 | 69.78±0.19 |
| SVDD [12] | 72.91±0.55 | 64.9±1.4 | 79.21±0.19 | 56.71±0.62 | 60.64±1.23 | 69.75±0.21 |
| KOC [20] | 65.03±1.1 | 66.39±0.53 | 79.04±0.33 | 54.78±0.21 | 57.09±1.48 | 68.81±0.99 |
| AEKOC [21] | 67.99±0.94 | 61.15±0.68 | 78.66±0.28 | 54.01±0.39 | 56.19±0.68 | 68.31±0.53 |
| GKOC-LDA [22] | 64.44±0.92 | 62.88±0.8 | 78.94±0.47 | 54.57±0.3 | 57.04±1.37 | 68.81±1.04 |
| LKOC-LE [22] | 64.19±0.96 | 64.38±0.78 | 79.02±0.33 | 54.58±0.26 | 57.12±1.66 | 68.77±0.92 |
| LKOC-LLE [22] | 59±1.14 | 66±0.44 | 77.5±0.68 | 52.02±0.56 | 56.28±0.79 | 67.72±0.47 |
| GKOC-CDA [22] | 64.39±0.74 | 64.84±0.94 | 78.89±0.48 | 54.5±0.33 | 57.07±1.38 | 68.81±1.04 |
| GKOC-CV[23] | 69.02±0.67 | 65.66±0.4 | 77.56±0.14 | **58.91**±0.4 | 62.78±0.56 | **74.42**±0.9 |
| GKOC-SV [23] | 68.23±2.18 | 67.28±0.75 | 78.36±0.37 | 55.59±0.77 | 58.85±1.45 | 70.07±2.48 |
| MKOC_$\theta$1 | 71.72±1.01 | **68.89**±0.5 | 79.21±0.5 | 57.7±0.86 | 62.55±0.37 | 73.56±0.49 |
| MKOC_$\theta$2 | **73.93**±1.46 | 59.51±1.88 | **80.5**±0.3 | 57.8±0.28 | 62.19±0.79 | 73.61±0.87 |

Table 4: Performance in term of $\eta_g \pm \Delta$ (%) over 5-folds and 5 runs for three miscellaneous datasets

| One-class Classifier | Glass(1) | Glass(2) | Iris |
| --- | --- | --- | --- |
| KPCA [14] | 57.69±0.71 | 77.65±0.21 | 96.44±0.61 |
| OCSVM [13] | 59.61±0.55 | 73.32±0.88 | 85.06±2.6 |
| SVDD [12] | 59.61±0.55 | 72.89±0.69 | 84.12±2.86 |
| KOC [20] | 58.91±1.18 | 73.08±0.77 | 92.35±0.87 |
| AEKOC [21] | 59.29±1.03 | 73.22±0.77 | 92.79±0.91 |
| GKOC-LDA [22] | 58.55±1.73 | 73.14±0.56 | 89.67±0.63 |
| LKOC-LE [22] | 59.01±1.37 | 72.94±0.44 | 90.35±0.07 |
| LKOC-LLE [22] | 59.23±1.74 | 72.07±1.3 | 92.35±0.87 |
| GKOC-CDA [22] | 58.49±1.75 | 73.14±0.56 | 89.67±0.63 |
| GKOC-CV[23] | 56.7±1.41 | **79**±1.54 | 94.69±1.01 |
| GKOC-SV [23] | 61.28±3.36 | 74.43±0.93 | 94.97±2.53 |
| MKOC_$\theta$1 | **62.46**±1.51 | 77.15±0.47 | **99.59**±0.56 |
| MKOC_$\theta$2 | 59.23±0.96 | 75.69±2.09 | 93.87±2.97 |

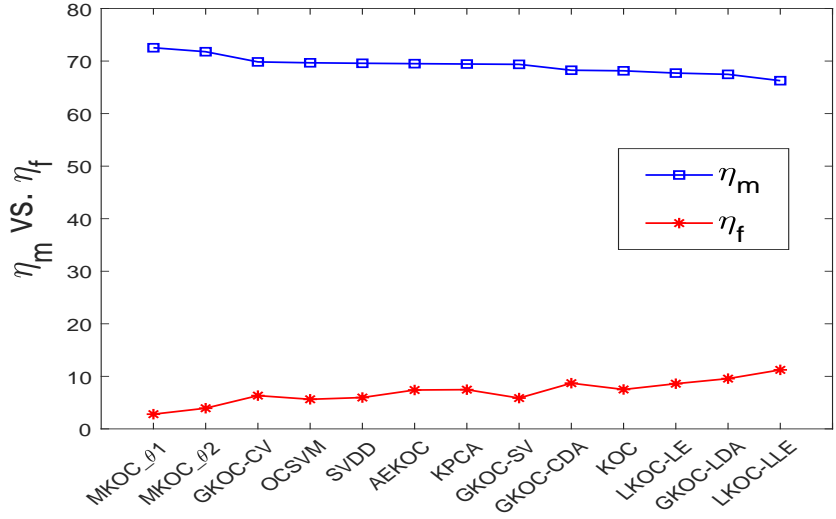

Figure 2: All one-class classifiers as per $\eta_m$ in decreasing order and their corresponding Friedman Rank ($\eta_f$).

Table 5: $\eta_f$ and $\eta_m$ of all one-class classifiers in increasing order of the $\eta_f$ (less value of $\eta_f$ indicates better performance).

| One-class Classifier | $\eta_f$ | $\eta_m(\%)$ |
|---|---|---|
| MKOC_$\theta$1 | 2.80 | 72.53 |
| MKOC_$\theta$2 | 3.93 | 71.78 |
| OCSVM | 5.63 | 69.67 |
| GKOC-SV | 5.87 | 69.37 |
| SVDD | 5.97 | 69.58 |
| GKOC-CV | 6.33 | 69.84 |
| AEKOC | 7.40 | 69.50 |
| KPCA | 7.47 | 69.44 |
| KOC | 7.50 | 68.14 |
| LKOC-LE | 8.60 | 67.71 |
| GKOC-CDA | 8.70 | 68.26 |
| GKOC-LDA | 9.57 | 67.47 |
| LKOC-LLE | 11.23 | 66.27 |

Especially, for Iris dataset, *MKOC_$\theta$1* exhibits significant improvement (>4%) of $\eta_g$ compared to all of the methods presented in Table 4. In case of Glass(2) dataset, *MKOC_$\theta$1* yields better result compared to all of the methods presented in Table 4 except *GKOC-CV* and *KPCA*. As we have discussed earlier that 7 datasets are imbalanced in nature, namely, German(1), German(2), Pima(1), Pima(2), Glass(1), Glass(2), and Iris. Out of these 7 datasets, proposed multi-layer one-class classifier performs better for 5 datasets. Moreover, *MKOC_$\theta$1* and *MKOC_$\theta$2* yield better results for 3 and 2 datasets, respectively.

Overall, it can be observed from the above discussion and Tables 2 to 4 that *MKOC_$\theta$2*, *MKOC_$\theta$1*, *GKOC-CV*, *AEKOC*, and *KPCA* yield best results for 5, 4, 3, 2 and 1 datasets, respectively. Further, we compute $\eta_m$ and $\eta_p$ for all the classifiers to analyze the $\eta_g$ value more closely.

The performance of each method over the 15 datasets using the $\eta_m$ metric is presented in Table 5 and is plotted in a decreasing order in Fig. 2. Based on the obtained results in Fig. 2, it can be clearly stated that both proposed variants of *MKOC*, i.e., *MKOC_$\theta$1*, *MKOC_$\theta$2* have achieved top two positions among 13 one-class classifiers as per $\eta_m$ criterion. However, *GKOC-CV* yields best $\eta_m$ among existing kernel-based one-class classifiers. It is to be noted that *MKOC_$\theta$2* yields best $\eta_g$ for maximum number (i.e., 5) of datasets, however, *MKOC_$\theta$1* yields better $\eta_m$ compared to *MKOC_$\theta$2*. Hence, in order to further analyze the performance of the competing one-class classifiers, $\eta_p$ is calculated as per Eq. (31), similar to [40].

$\eta_p$ metric provides information regarding proximateness of each classifier towards maximum $\eta_g$ value. As it can be seen in Table 6, *MKOC_$\theta$1* and *MKOC_$\theta$2* hold the top two positions similar to the ranking based on the $\eta_m$ values in Fig. 2. In Fig. 3, $\eta_p$ values of all 13 one-class classifiers are plotted in an increasing order for all of the datasets. *MKOC_$\theta$1* is a multi-layer version of the single-layered classifier *KOC*. The plotted lines for these two classifiers in Fig. 3 clearly indicate the performance improvement of multi-layer version over single-layer one. Overall, Fig. 3 illustrates the clear superiority of the multi-layer one-class classifiers over the existing methods. Moreover, $MKOC\_\theta1$ obtains more than 97% $\eta_p$ value for all datasets except German(1) dataset. Detailed $\eta_p$ values for all 13 classifiers over 15 datasets are made available on the web page (https://goo.gl/XLBtqp).

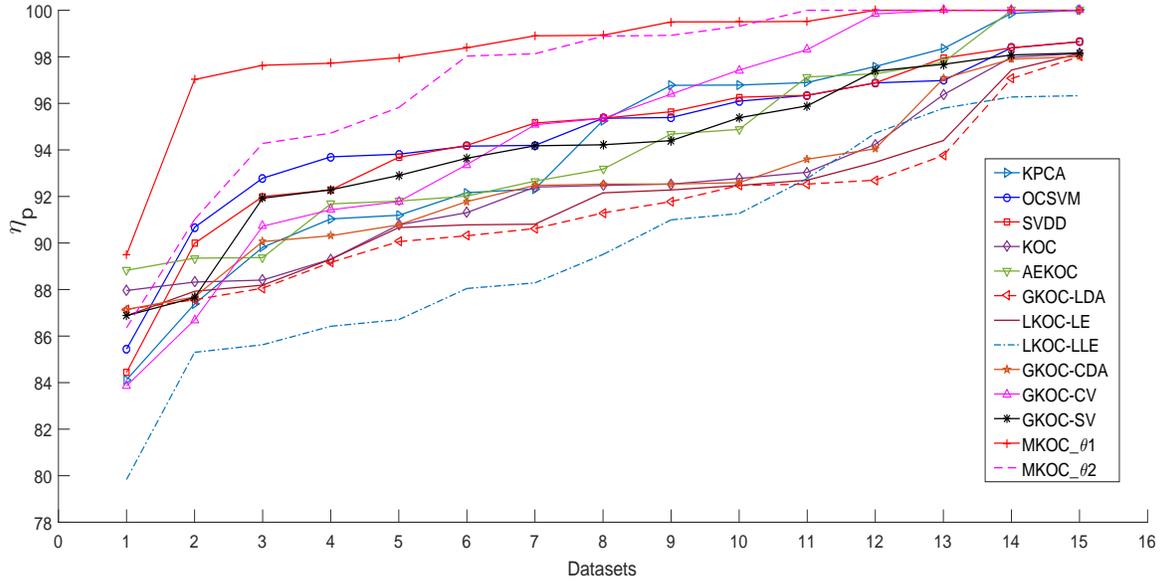

Figure 3: $\eta_p$ achieved by various one-class classifiers over 15 datasets (ordered by increasing percentage)

Table 6: $\eta_p$ value over 15 datasets

| One-class Classifiers | $\eta_p$ (%) |
|---|---|
| MKOC_$\theta$1 | 98.31 |
| MKOC_$\theta$2 | 97.03 |
| GKOC-CV | 94.68 |
| OCSVM | 94.59 |
| SVDD | 94.48 |
| GKOC-SV | 94.04 |
| AEKOC | 94.04 |
| KPCA | 93.97 |
| GKOC-CDA | 92.57 |
| KOC | 92.40 |
| LKOC-LE | 91.84 |
| GKOC-LDA | 91.50 |
| LKOC-LLE | 89.86 |

Above discussion suggests $MKOC\_\theta1$ and $MKOC\_\theta2$ as the best performing classifier in term of $\eta_g$, $\eta_m$, and $\eta_p$. Despite this fact, a statistical testing needs to perform for verifying this fact. In the next subsection, Friedman Rank ($\eta_f$) testing is performed for statistical testing.

*3.3. Statistical Comparison*

For comparing the performance of the proposed variants $MKOC\_\theta1$ and $MKOC\_\theta2$ with the 11 existing kernel-based methods on 15 benchmark datasets, a non-parametric Friedman test is employed. In the Friedman test, the null hypothesis states that the mean of individual experimental treatment is not significantly different from the aggregate mean across all treatments and the alternate hypothesis states the other way around. Friedman test mainly computes three components viz., F-score, p-value and Friedman Rank ($\eta_f$). If the computed F-score is greater than the critical value at the tolerance level $\alpha = 0.05$, then one rejects the equality of mean hypothesis (i.e. null hypothesis). We employ the modified Friedman test [35] for the testing, which was proposed by Iman and Davenport [41]. The F-score obtained after employing non-parametric Friedman test is 7.28, which is greater than the critical value at the tolerance level $\alpha = 0.05$ i.e. $7.28 > 1.81$. Hence, null hypothesis can be rejected with 95% of a confidence level. The computed p-value of the Friedman test is $1.1461e - 08$ with the tolerance value $\alpha = 0.05$, which is much lower than 0.05. This small value indicates that differences in the performance of the various methods are statistically significant.

Afterwards, $\eta_f$ of each classifier is also calculated to assign a rank to all 13 one-class classifiers. Friedman test assigns a rank to all the methods for each dataset, it assigns rank 1 to the best performing algorithm, the second best rank 2 and so on. If rank ties then average ranks are assigned [35]. The $\eta_f$ value of all classifiers is provided in increasing order of its value (less value of $\eta_f$ indicates better performance) in Table 5. These values are visualized in Fig. 2 with the decreasing order of $\eta_m$. $MKOC\_\theta1$ and $MKOC\_\theta2$ still achieve top two positions, similar to using the $\eta_m$ metric. From Table 5 and Fig. 2, it can be observed that $\eta_f$ of most of the classifiers follows a similar pattern as $\eta_m$, i.e., $\eta_f$ increases as $\eta_m$ decreases. However, some of the one-class classifiers don't follow the same pattern as with $\eta_m$ like *GKOC-CV* which has better $\eta_m$ but inferior $\eta_f$ compared to *OCSVM* and *GKOC-SV*. The above analysis indicates that an one-class classifier with better $\eta_f$ has better generalization capability compared to the other existing methods.

Overall, after the performance analysis of all the 13 one-class classifiers, it is observed that none of the existing one-class classifiers perform better than the proposed multi-layer one-class classifier in term various above discussed performance criteria.

## 4. Conclusion

This paper has presented two variants of *KRR*-based multi-layer one-class classifier. It is constructed by stacking various Auto-Encoders followed by a *KRR*-based one-class classifier. Hence, two types of training processes are involved i.e. one is for *KAE* and other is for *KOC*. Extensive experimental comparisons have been provided with 11 state-of-the-art kernel feature mapping based one-class classifiers over 15 publicly available datasets in term of $\eta_g$, $\eta_m$, $\eta_p$, and $\eta_f$. These experiments have exhibited that the proposed multi-layer one-class classifier provides state-of-the-art performance. Stacked Auto-Encoder through multiple layers helps *MKOC* in achieving better generalization and data representation capability. Moreover, the statistical significance of the results has been verified by Friedman Ranking test. In future work, various other types of available Auto-Encoder can be explored to enhance the performance of the proposed multi-layer architecture.

## Acknowledgment

This research was supported by Department of Electronics and Information Technology (DeITY, Govt. of India) under Visvesvaraya PhD scheme for electronics & IT.